\documentclass[letterpaper, 10 pt, conference]{ieeeconf}  % Comment this line out
\usepackage{multirow}
\usepackage{amssymb,graphicx,amsmath,color,algorithm,algorithmic}
\usepackage[noadjust]{cite} % groups citations
\usepackage{caption}
\usepackage{subcaption}
\usepackage{eucal}
\usepackage{layouts}
\graphicspath{{figures/}}

\IEEEoverridecommandlockouts                              % This command is only
\title{\LARGE \bf
Emergence of Human-comparable Balancing Behaviors by Deep Reinforcement Learning
\vspace{0em}
}

\author{Chuanyu Yang, Taku Komura, Zhibin Li% <-this % stops a space
	\thanks{Chuanyu Yang, Taku Komura,  and Zhibin Li are with the School of Informatics, The University of Edinburgh, UK. \newline Email: \{chuanyu.yang, tkomura, zhibin.li\}@ed.ac.uk}
}

\begin{document}

\maketitle
\thispagestyle{empty}
\pagestyle{empty}

% keywords: 	Force and Tactile Sensing 	Biomimetics

%%%%%%%%%%%%%%%%%%%%%%%%%%%%%%%%%%%%%%%%%%%%%%%%%%%%%%%%%%%%%%%%%%%%%%%%%%%%%%%%
\begin{abstract}
This paper presents a hierarchical framework based on deep reinforcement learning that learns a diversity of policies for humanoid balance control. Conventional zero moment point based controllers perform limited actions during under-actuation, whereas the proposed framework can perform human-like balancing behaviors such as active push-off of ankles. The learning is done through the design of an explainable reward based on physical constraints. The simulated results are presented and analyzed. The successful emergence of human-like behaviors through deep reinforcement learning proves the feasibility of using an AI-based approach for learning humanoid balancing control in a unified framework.% The operating range of the balance strategy learnt by reinforcement learning is larger than that of conventional zero moment point based approaches, revealing the potential of using reinforcement learning to learn control strategies to replace engineering based strategies.

\vspace{-0em}
\end{abstract}

%%%%%%%%%%%%%%%%%%%%%%%%%%%%%%%%%%%%%%%%%%%%%%%%%%%%%%%%%%%%%%%%%%%%%%%%%%%%%%%%
\section{Introduction}
\label{sec:1}
Humans efficiently make use of under-actuated control, such as toe tilting and heel rolling, for keeping balance when standing and walking. Biomechanical study of human walking has discussed about the advantage of rolling around the heel and toe during walking phase \cite{cite:adamczyk2006AdvantagesRollingFoot}. From a biomechanical point of view, tilting the foot creates better foot-ground clearance allowing the maximum ankle torques to be exploited \cite{cite:li2015ActiveControl}, \cite{ cite:li2017HumanoidBalancing}.

Foot tilting give rise to a control problem as an underactuated degree of freedom (DOF) is introduced. Once foot tilting occurs, the edge of the foot namely the heel or toe, becomes the sole contact point between the foot and the ground which the body pivots around. This new pivot point is an underactuated DOF as zero torque can be applied on the pivoting axis. The physically feasible range of center of pressure (COP) converges to a singular boundary line on the edge of the foot. The controller has no control authority over the new underactuated DOF since no torque can be applied.

Many modern humanoid robots are designed to closely resemble the human morphology. In theory, they possess similar capabilities of a human, and should be able to perform foot tilting behaviors comparable to humans. However, most robots are shown to focus on keeping the foot flat on the ground during locomotion, which is unnatural and inefficient. The reason is not due to the physical capabilities, but rather because of the limitation of modern control and actuation techniques. Most balance and walking control approaches are based on ZMP, and are developed on the assumption of a fully actuated system where the foot is placed flat on the ground creating a large size of support polygon. Most ZMP based methods will fail during underactuation phases, as they need to restrict the ZMP or COP to be within the support polygon created by a flat foot and away from the foot edges.

Controllers that permits the COP to lie on the narrow boundary of the foot have been developed to generate underactuated foot tilting behaviors, demonstrating the feasibility of using analytic engineering approach for designing controllers capable of dealing with underactuated phases during balance recovery \cite{cite:li2017HumanoidBalancing}.

\begin{figure}[t]
	\centering
	\vspace{0em}
	%\fbox{
	{\includegraphics[width=0.7\linewidth, trim = 2.8cm 8.8cm 1.6cm 2.8cm, clip]{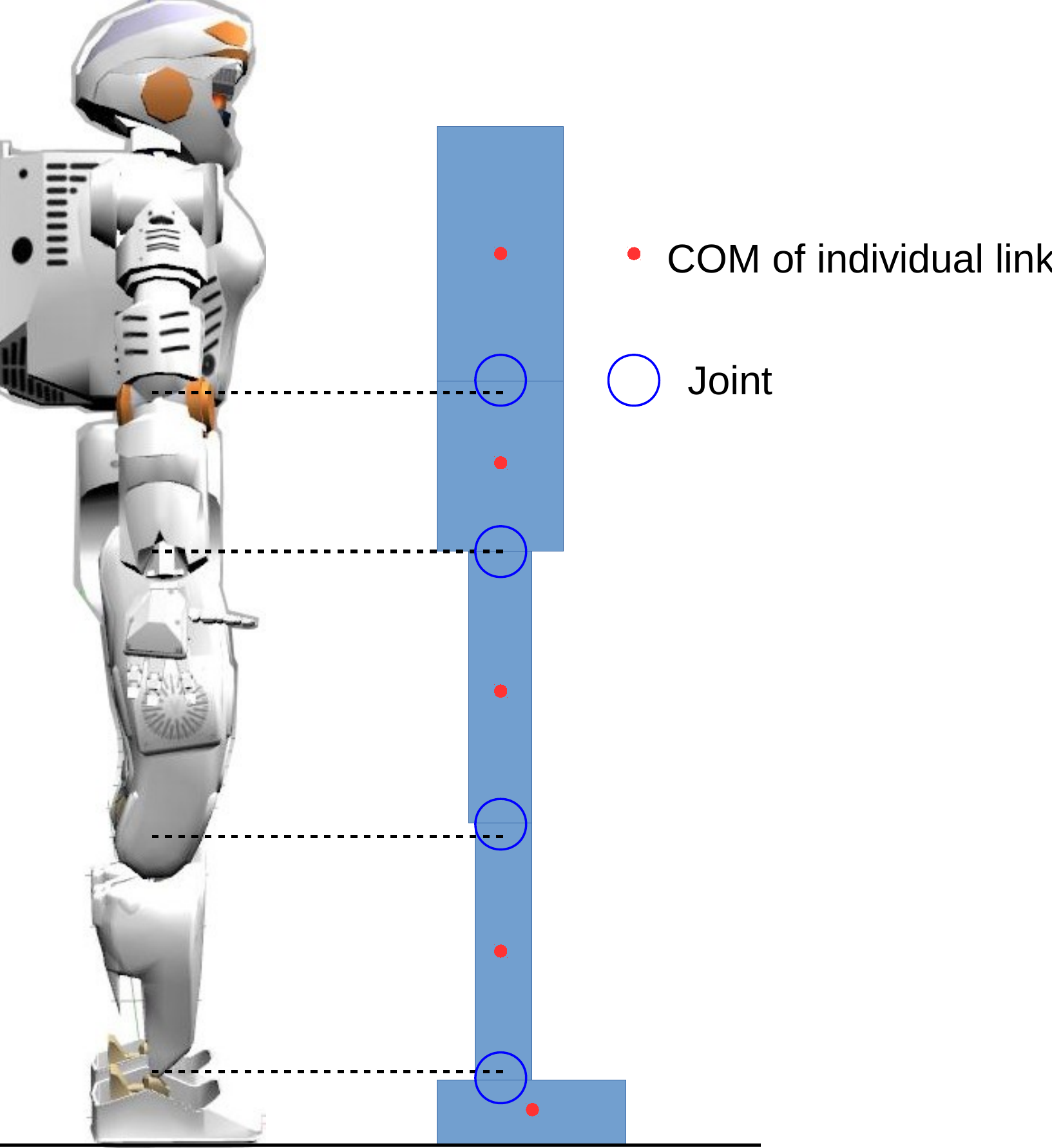}}
	%}\\
	\caption{Side view of valkyrie robot and the 2D humanoid character modelled according to Valkyrie robot.}
	\label{fig:1}
	\vspace{0em}
\end{figure}

Recently, machine learning approaches such as deep reinforcement learning have been attracting robotics researchers, which can automatically learn the parameters for achieving the given objective. Engineering based approaches require a lot of human knowledge in designing the controllers and additional effort in tuning, which is a disadvantage. Machine learning approaches such as deep reinforcement learning (RL) have a major advantage: they require less manual tuning. Certainly, reinforcement learning also requires a certain amount of human knowledge and effort while designing the RL agent and the reward, but rather the main effort is in the knowledge-based construction of the agent and reward, instead of structuring explicit controllers. After the proper RL agent and reward are constructed, the agent is capable of learning the optimal policy by itself. Recent works done on deep reinforcement learning have demonstrated that deep RL is capable of dealing with very complex and dynamic motor tasks in continuous state and action spaces. Therefore, it is clear that deep reinforcement learning should have the capability of learning a policy to deal with both flat foot and foot tilting situations.

In this paper, we propose a novel framework based on deep reinforcement learning that can make use of under-actuated behavior for keeping the balance during standing. Since deep reinforcement learning paradigm is shown to allow very distinct and complex behaviours to emerge from simple rewards \cite{cite:heess2017EmergenceOf}, following the prior work \cite{cite:li2017HumanoidBalancing}, we are motivated to explore an alternative to use deep reinforcement learning to acquire a policy that is capable of generating human-comparable behaviors during push recovery without providing any explicit knowledge of the control policies.

The contributions of our study are the following:
\begin{itemize}
	\item Provided a physical analysis on the reward design of humanoid balancing;
	\item Demonstrated that deep reinforcement learning is capable of learning a human like balancing strategy with limited knowledge provided.
\end{itemize}

 %We will demonstrate that equivalent performances can also be generated using machine learning approaches like deep reinforcement learning.

%%%%%%%%%%%%%%%%%%%%%%%%%%%%%%%%%%%%%%%%%%%%%%%%%%%%%%%%%%%%%%%%%%%%%%%%%%%%%%%%

\section{Related Work and Motivation}
\label{sec:2}

Recent breakthroughs in reinforcement learning and deep learning, have given rise to deep reinforcement learning (deep RL), which is a combination of reinforcement learning and deep neural networks. The rise of deep RL has enhanced the capability of agents to perform more complex and dynamic tasks in high dimensional continuous state and action spaces. There are quite a few well known deep RL algorithms dedicated to solving problems in continuous state and action spaces such as Trust Region Policy Optimization (TRPO) \cite{cite:schulman2015TrustRegion}, Normalized Advantage Function (NAF) \cite{cite:gu2016ContinuousDeepQ}, Asynchronous Advantage Actor Critic (A3C) \cite{cite:mnih2016AsynchronousMethods} and Deep Deterministic Policy Gradient (DDPG) method \cite{cite:lillicrap2016ContinuousControl}.

Researchers in the computer science and robotics community have published a few papers on using deep reinforcement learning for humanoid motion control. Peng et al. successfully applied Continuous Actor Critic Learning Automaton (CACLA) \cite{cite:van2007ReinforcementLearning}, \cite{cite:van2012ReinforcementLearning} to train a bipedal character to learn terrain traversal skills for terrain with gaps and walls \cite{cite:peng2015DynamicTerrain}. Later, they developed a hierarchical deep reinforcement learning framework, having the low-level controller (LLC) to specialize on balance and limb control, while the high-level controller (HLC) focuses on navigation and trajectory planning. Using their framework, the bipedal character successfully learned the skills in
order to perform tasks such as soccer ball guiding, path following and obstacle avoidance \cite{cite:peng2017DeepLoco}.

Kumar et al. used deep reinforcement learning to learn a safe falling strategy for humanoids to minimize damage during fall. Their algorithm is based on CACLA and the Mixture of Actor-Critic Experts (MACE) architecture \cite{cite:peng2016TerrainAdaptive}. In this architecture, each joint is assigned with an independent actor-critic pair. The actor with the highest corresponding critic value will be activated to generate action. This architecture combines both continuous control and discrete controls.

A lot of control methods have been proposed for humanoid balancing \cite{cite:hyon2009IntegrationOf}, \cite{cite:stephens2010DynamicBalance}, \cite{cite:li2012StabilizationFor}. However, most controllers do not deal with foot tilting during humanoid locomotion as it restricts the center of pressure to a single point causing the system to be underactuated creating immense difficulties for the design of controllers. Instead, this problem is bypassed by restricting the foot to remain flat on the ground. However, this flat foot balancing behaviour is different from what we humans do.

Few works have been done on the topic of balance recovery by active foot tilting. Li et al. have done a thorough analysis on the dynamics of foot tilting and derived a foot tilting (ankle push-off) balancing strategy. They explained the underlying mechanism and the significance of foot tilting \cite{ cite:li2017HumanoidBalancing}. Concrete physical and mathematical proof in \cite{ cite:li2017HumanoidBalancing} suggest that foot tilting balance strategy is more robust against force perturbations than flat foot balance strategy, and have successfully designed a controller capable of underactuated foot tilting and implemented it on a real robot.

Since the physical viability of stable underactuated behaviours has been achieved in a deterministic and analytic approach using modern control techniques \cite{ cite:li2017HumanoidBalancing}, our study hereby aim to answer whether similar human-like behavior can be a natural outcome using machine learning approach, specifically, deep reinforcement learning. Our task has to be performed in a continuous environment, thus the deep reinforcement learning has to be applicable in continuous state and action spaces.

%%%%%%%%%%%%%%%%%%%%%%%%%%%%%%%%%%%%%%%%%%%%%%%%%%%%%%%%%%%%%%%%%%%%%%%%%%%%%%%%
\section{Preliminaries}
\label{sec:3}

In this section we briefly review some of the concepts essential for understanding the paper.

\subsection{State Representation}
\label{sec:3a}

The bipedal character configuration used in this paper is shown in Fig. \ref{fig:1}, it is roughly modelled according to Valkyrie, with the notion to apply the deep RL humanoid balancing strategy on the Valkyrie robot in the future.

\begin{figure}[t]
	\centering
	\vspace{0em}
	%\fbox{
	{\includegraphics[width=0.6\linewidth, trim = 5.3cm 13.3cm 5.9cm 5cm, clip]{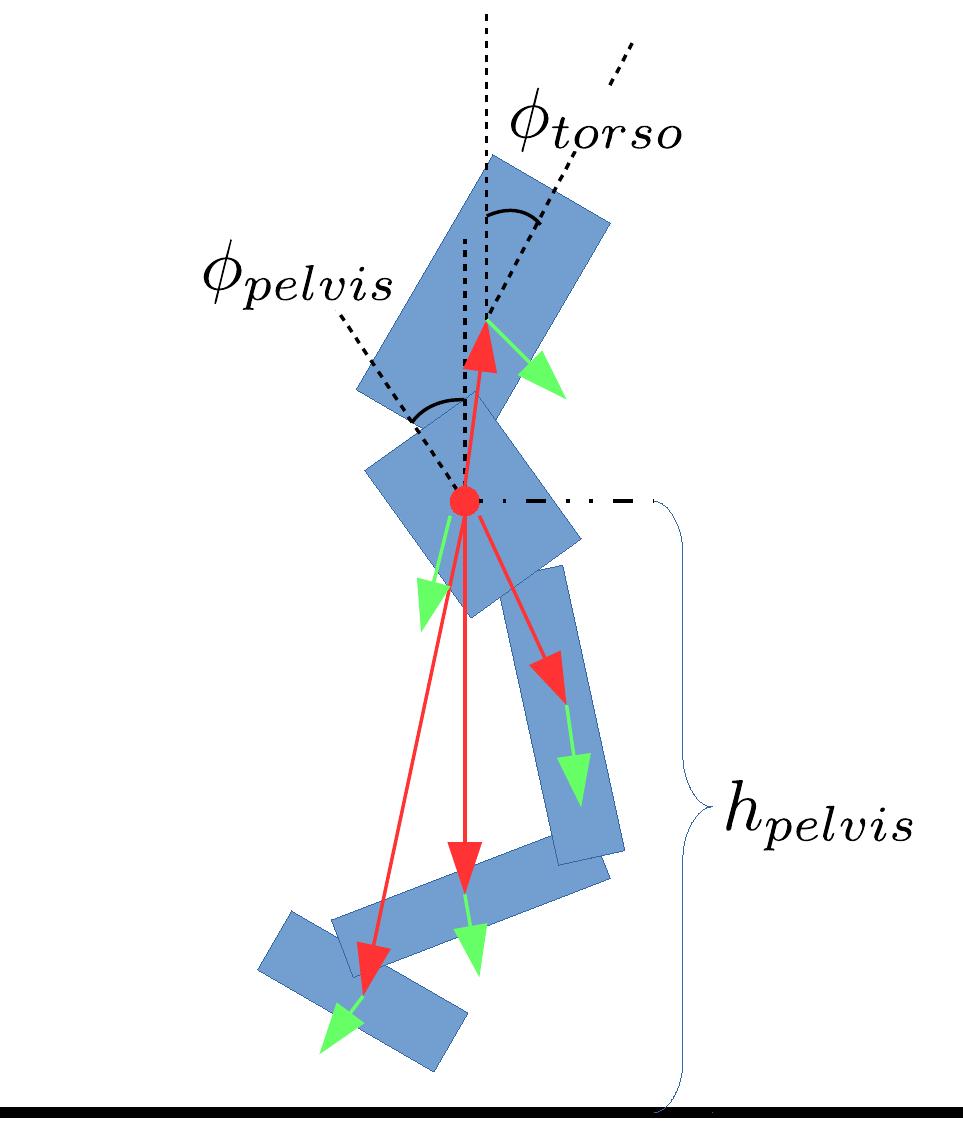}}
	%}\\
	\caption{State features for the biped.} % write a better caption that reflects the meaning
	\label{fig:2}
	\vspace{0em}
\end{figure}

The state input for reinforcement learning is crucial, the input state features should contain adequate amount of information that allows the reinforcement learning algorithm to learn a good policy, but it should not be too redundant as increasing the input dimension increases training time and introducing irrelevant parameters can hinder the performance. In our case, the state should contain enough information about the kinematics and the dynamics of the humanoid.

Fig. \ref{fig:2} shows the selected state features, it includes pelvis height ($h_{\text{height}}$), joint angle and joint velocity, the angle ($\phi_{\text{torso}}$, $\phi_{\text{pelvis}}$) and angular velocity($\dot{\phi}_{\text{torso}}$, $\dot{\phi}_{\text{pelvis}}$) of the pelvis and torso, ground contact information, displacement of COM of links with reference to (w.r.t) the pelvis (red) and the linear velocities of all body links (green).

\subsection{Capture Point}
\label{sec:3b}
Capture point is a concept commonly used in humanoid locomotion, it is defined {as a point on the ground where the robot can step to in order to bring itself to a complete stop \cite{cite:pratt2006CapturePoint}. Knowing the velocity and height of the inverted pendulum, and the gravitational acceleration, we are able to compute the capture point,
\begin{align}
x_{\text{capture}}&=x_{\text{\tiny{COM}}}+\dot{x}_{\text{\tiny{COM}}}\sqrt{\frac{z_{0}}{g}} ,
\label{eq:1}\\
%\dot{x}=\sqrt{\frac{g}{z_{c}}}\delta_{cop}
J_{\text{reject}}&=m\sqrt{\frac{g}{z_{c}}}\Delta_{\text{COP}} ,
\label{eq:2}
\end{align}
where the impulse $J_{\text{reject}}$ derived by capture point is the theoretical maximum of the impulse that can be rejected, where $z_{c}$ is the COM height, $\Delta_{\text{COP}}$ is the relative horizontal distance between the constant COP and the initial COM position, $m$ is the total mass of the inverted pendulum \cite{cite:li2017HumanoidBalancing}.

%Normally during humanoid balancing, the center of mass and center of pressure are within the support polygon created by the foot.
The capture point as an indication of balance is considered to be within the support polygon, so the maximum reachability of the capture point is at the edge of the foot. If the impulse time $t_{\text{impulse}}$ and the mass of the humanoid $m$ are known, we can derive other useful physical properties such as the maximum force $F_{\text{max}}$ the robot can withstand and the maximum velocity disturbance $V_{\text{max}}$ of the COM when the robot is still able to balance.
\begin{align}
V_{\text{max}}&=\frac{J_{\text{reject}}}{m}
\label{eq:3} \\
F_{\text{max}}&=\frac{J_{\text{reject}}}{t_{\text{impulse}}}
\label{eq:4}
\end{align}

\subsection{Explainable Design of a Reward}
\label{sec:3c}

\begin{figure}[t]
	\centering
	%\fbox{
		{\includegraphics[width=0.95\linewidth, trim = 1.2cm 11.9cm 1.1cm 5.5cm, clip]{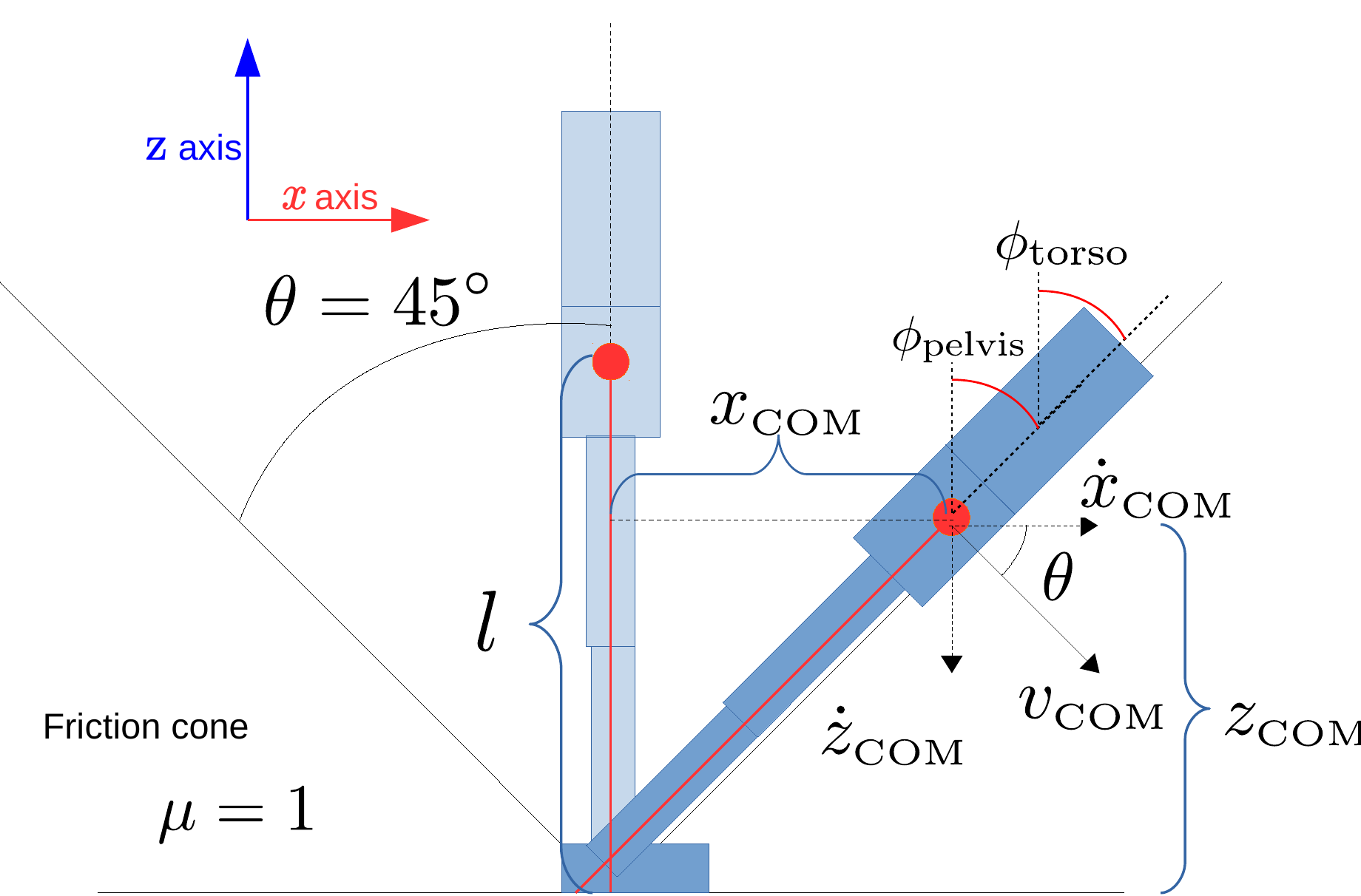}}
	%}\\
	\vspace{0em}
	\caption{$\dot{x}_{\text{\tiny{COM}}}$ and $\dot{z}_{\text{\tiny{COM}}}$ are the actual COM velocities while the humanoid falls under the force of gravity. $x_{\text{\tiny{COM}}}$ and $z_{\text{\tiny{COM}}}$ is the position of the COM on the sagittal plane. $\phi_{\text{torso}}$ and $\phi_{\text{torso}}$ are orientation of the upper body. $l$ is the length from the center of the foot to the COM.}
	\label{fig:3}
	\vspace{0em}
\end{figure}

In this section, we describe the design of our reward function for the deep reinforcement learning for maintaining the balance of our humanoid model, which is based on ideas such as capture point and linear inverted pendulum model from conventional control techniques.

The reward has to be designed carefully in order to produce desired results: Amodei et al. has mentioned the performance and safety issues that can be caused by a poorly designed reward function in reinforcement learning \cite{cite:amodei2016ConcreteProblems}. 
To deal with the difficulty of designing a functional reward with human knowledge, researchers came up with methods such as inverse reinforcement learning, sometimes referred as imitation learning or apprenticeship learning. The idea is to elicit the reward function from a given sequence of demonstrations \cite{cite:ng2000AlgorithmsFor,cite:abbeel2004ApprenticeshipLearning,cite:christiano2017DeepReinforcement}.
In our case, we have no knowledge on how the humanoid should react in order to produce balancing behaviou that involves foot tiltingr. Therefore inverse reinforcement learning is not an option, as we are not able to provide proper demonstration.

%The reward has to be designed carefully, different objectives has to be %provided with different weights depending on their contribution to the tasks. %In the future, we could apply a more intelligent self adaptive multi-objective %reward that tunes the weight of those different objectives according to %performance.

Balancing can be decomposed into six objectives, turning the task into a multi-objective problem: 
keeping the torso and pelvis orientation upright (denoted by $r_{\phi_{\text{torso}}}$ and $r_{\phi_{\text{pelvis}}}$ in the reward), keeping the horizontal position of the COM close to the center of the foot (denoted by $r_{x_{\text{\tiny{COM}}}}$), keeping the COM vertical position at a certain height (denoted by $r_{z_{\text{\tiny{COM}}}}$) and minimizing the horizontal and vertical velocity of the COM (denoted by $r_{\dot{x}_{\text{\tiny{COM}}}}$ and $r_{\dot{z}_{\text{\tiny{COM}}}}$).
The total reward is the linear combination of the individual objectives as:
\begin{equation}
\begin{aligned}
	r  &= w_{\phi_{\text{torso}}}r_{\phi_{\text{torso}}}+ w_{\phi_{\text{pelvis}}}r_{\phi_{\text{pelvis}}}+
w_{x_{\text{\tiny{COM}}}}r_{x_{\text{\tiny{COM}}}}\\
 	& +w_{z_{\text{\tiny{COM}}}}r_{z_{\text{\tiny{COM}}}}+
w_{\dot{x}_{\text{\tiny{COM}}}}r_{\dot{x}_{\text{\tiny{COM}}}}+
w_{\dot{z}_{\text{\tiny{COM}}}}r_{\dot{z}_{\text{\tiny{COM}}}},
\label{eq:5}
\end{aligned}
\end{equation}
where, the weights $w_{[\cdot]}$ of each objective in the reward are set to 1 in default except  
 $w_{z_{\text{\tiny{COM}}}}$ which is set to 5 for counteracting gravity.

The individual reward terms $r_{[\cdot]}$ are defined such that the individual parameters are attracted to their target values,
while degrading exponentially as getting farther:  
\begin{equation}
\begin{aligned}
&r_{\phi_{\text{torso}}}=\exp(-\alpha_{\phi_{\text{torso}}}~\phi_{\text{torso}}^2)\\
&r_{\phi_{\text{pelvis}}}=\exp(-\alpha_{\phi_{\text{pelvis}}}~\phi_{\text{pelvis}}^2)\\
&r_{x_{\text{\tiny{COM}}}}=\exp(-\alpha_{x_{\text{\tiny{COM}}}}~(x_{\text{\tiny{COM}}}^{\text{target}}-x_{\text{\tiny{COM}}})^2)\\
&r_{z_{\text{\tiny{COM}}}}=\exp(-\alpha_{z_{\text{\tiny{COM}}}}~(z_{\text{\tiny{COM}}}^{\text{target}}-z_{\text{\tiny{COM}}})^2)\\
&r_{\dot{x}_{\text{\tiny{COM}}}}=\exp(-\alpha_{\dot{x}_{\text{\tiny{COM}}}}~(\dot{x}_{\text{\tiny{COM}}}^{\text{target}}-\dot{x}_{\text{\tiny{COM}}})^2)\\
&r_{\dot{z}_{\text{\tiny{COM}}}}=\exp(-\alpha_{\dot{z}_{\text{\tiny{COM}}}}~(\dot{z}_{\text{\tiny{COM}}}^{\text{target}}-\dot{z}_{\text{\tiny{COM}}})^2) .\\
\end{aligned}
\phantom{\hspace{0cm}}
\label{eq:6}
\end{equation}
The target for the orientation of the torso and pelvis is 0 rad. $x_{\text{\tiny{COM}}}^{\text{target}}$ and $z_{\text{\tiny{COM}}}^{\text{target}}$ are the target horizontal and vertical COM position. $\dot{x}_{\text{\tiny{COM}}}^{\text{target}}$ and $\dot{z}_{\text{\tiny{COM}}}^{\text{target}}$ are the target horizontal and vertical COM velocity.

A preliminary normalization factor $\alpha_{[\cdot]}$ is introduced because the the units and range of the values of the physical properties are different. For normalization, we expect the individual rewards presented in (\ref{eq:6}) are lower than $\epsilon = 1\times10^{-5}$ when it reaches maximum error range. By obtaining the maximum error range $e_{[\cdot]}$ of each physical properties, we can calculate the normalization factor $\alpha_{[\cdot]}$  that meets the requirement as

\begin{equation}
\begin{aligned}
&\alpha_{\phi_{\text{torso}}} = -\ln(\epsilon)/e_{\phi_{\text{torso}}}^2\\
&\alpha_{\phi_{\text{pelvis}}} = -\ln(\epsilon)/e_{\phi_{\text{pelvis}}}^2\\
&\alpha_{x_{\text{\tiny{COM}}}} = -\ln(\epsilon)/e_{x_{\text{\tiny{COM}}}}^2\\
&\alpha_{z_{\text{\tiny{COM}}}} = -\ln(\epsilon)/e_{z_{\text{\tiny{COM}}}}^2\\
&\alpha_{\dot{x}_{\text{\tiny{COM}}}} = -\ln(\epsilon)/e_{\dot{x}_{\text{\tiny{COM}}}}^2\\
&\alpha_{\dot{z}_{\text{\tiny{COM}}}} = -\ln(\epsilon)/e_{\dot{z}_{\text{\tiny{COM}}}}^2 .\\
\end{aligned}
\phantom{\hspace{0cm}}
\label{eq:7}
\end{equation}

We now describe about how to compute the maximum error range $e_{[\cdot]}$ for each term.
Regarding the maximum error range for the torso angle $e_{\phi_{\text{torso}}}$ and pelvis angle $e_{\phi_{\text{pelvis}}}$, that occurs when  $\phi_{\text{torso}} = \phi_{\text{pelvis}} = \pi/2$, which is when the body is fully horizontal. 
Thus, $e_{\phi_{\text{torso}}} = e_{\phi_{\text{pelvis}}} = \pi/2$ rad.

Regarding the maximum error range for the horizontal and vertical COM positions
$e_{x_{\text{\tiny{COM}}}}, e_{z_{\text{\tiny{COM}}}}$, we can approximate them by 
using the linear inverted pendulum model.
Fig. \ref{fig:3} shows the humanoid character lying on the boundary of the friction cone. We consider the situation in which the pendulum lies on the border of the friction cone as an extreme situation, any configuration that falls outside the friction cone is destined to fail and should be ignored. Assuming the coefficient of friction $\mu = 1$, the maximum angle of the friction cone $\theta_{\text{max}}$ is $\pi/4$ rad.
Under the situation shown in Fig. \ref{fig:3}, 
the maximum error range for the horizontal and vertical COM positions can be computed by
\begin{equation}
\begin{aligned}
e_{x_{\text{\tiny{COM}}}} &= -x_{_{\text{COM}}}\\
&= \sin(\theta_{\text{max}})l\\
\end{aligned}
\label{eq:8}
\end{equation}

\begin{equation}
\begin{aligned}
e_{z_{\text{\tiny{COM}}}} &= l-z_{_{\text{COM}}}\\
&= (1-\cos(\theta_{\text{max}}))l .
\end{aligned}
\label{eq:9}
\end{equation}
In our study, we use  $e_{x} = 0.768$ m and $e_{z}=0.318$ m.

Regarding the maximum error range for the horizontal and vertical COM velocities, 
$e_{\dot{x}_{\text{\tiny{COM}}}}, e_{\dot{z}_{\text{\tiny{COM}}}}$, 
we can compute them using the extreme orientation angle $\theta_{\text{max}}$ and the  
$V_{\text{max}}$ based on the capture point in \eqref{eq:3}. We consider the situation in which the pendulum lies on the boundary of the friction cone, where $\theta_{\text{max}} = \pi/4$ rad, to be the extreme condition. 
Given the orientation angle $\theta_{\text{max}}$,  
the height of the COM is $z_{0} =l\cos(\theta_{\text{max}})$ and the horizontal displacement of the COM to the COP is 
$\Delta_{\text{COP}} = -l \sin(\theta_{\text{max}})$. From \eqref{eq:2} and \eqref{eq:3}, we can compute 
$V_{\text{max}}=\sqrt{\frac{g}{z_{c}}}\Delta_{\text{COP}}$, which can be used to calculate 
the target horizontal velocity $\dot{x}_{\text{\tiny{COM}}}^{\text{target}}$. 
The target vertical velocity $\dot{z}_{\text{\tiny{COM}}}^{\text{target}}$ is set to 
0 as we wish to minimize the vertical movement of the COM. 
As a result, $e_{\dot{x}_{\text{\tiny{COM}}}}, e_{\dot{z}_{\text{\tiny{COM}}}}$ can be computed 
as follows:
\begin{equation}
\begin{aligned}
e_{\dot{x}_{\text{\tiny{COM}}}} &= \dot{x}_{\text{\tiny{COM}}}^{\text{target}} - \dot{x}_{\text{\tiny{COM}}}\\
&= -\sin(\theta_{\text{max}})l\sqrt{\frac{g}{\cos(\theta_{\text{max}})l}}\\
&~~~~-\cos(\theta_{\text{max}})\sqrt{2g(1-\cos(\theta_{\text{max}}))l}\\
\end{aligned}
\label{eq:10}
\end{equation}

\begin{equation}
\begin{aligned}
e_{\dot{z}_{\text{\tiny{COM}}}} &= \dot{z}_{\text{\tiny{COM}}}^{\text{target}}-\dot{z}_{\text{\tiny{COM}}}\\
&= -\sin(\theta_{\text{max}})\sqrt{2g(1-\cos(\theta_{\text{max}}))l}
\end{aligned}
\label{eq:11}
\end{equation}
The maximum error for horizontal and vertical COM velocity is thus 4.510 m/s and 1.766 m/s.% when $\theta = \pi/4$.

Applying the calculated maximum range of the physical properties back into \eqref{eq:7}, 
we obtain the normalization factor as
\begin{equation}
\begin{aligned}
&\alpha_{\phi_{\text{torso}}} = 4.67\\
&\alpha_{\phi_{\text{pelvis}}} = 4.67\\
&\alpha_{x_{\text{\tiny{COM}}}} = 19.50\\
&\alpha_{z_{\text{\tiny{COM}}}} = 113.74\\
&\alpha_{\dot{x}_{\text{\tiny{COM}}}} = 0.57\\
&\alpha_{\dot{z}_{\text{\tiny{COM}}}} =3.69 .\\
\end{aligned}
\phantom{\hspace{0cm}}
\label{eq:12}
\end{equation}
Using the normalization factors $\alpha_{[\cdot]}$,  
the individual reward components can be computed by \eqref{eq:6} 
and then the total reward by \eqref{eq:5}.

%\begin{equation}
%\begin{aligned}
%&w_{torso}=1\\
%&w_{pelvis}=1\\
%&w_{COM\_hor\_pos}=1\\
%&w_{COM\_ver\_pos}=5\\
%&w_{COM\_hor\_vel}=1\\
%&w_{COM\_ver\_vel}=1 .\\
%\end{aligned}
%\phantom{\hspace{0cm}}
%\label{eq:6}
%\end{equation}

\subsection{Deep deterministic policy gradient}
\label{sec:3d}
 
 \begin{algorithm}[t!]
 	\caption{Deep Deterministic Policy Gradient}
 	\label{alg:DDPG}
 	\begin{algorithmic} 
 		\STATE Initialize critic $Q(s,a|\theta^{Q})$ and actor network $\mu(s,a|\theta^{\mu})$
 		\STATE Initialize target networks $Q^{\prime}$ and $\mu^{\prime}$: $\theta^{Q^{\prime}}\leftarrow\theta^{Q}$,$\theta^{\mu^{\prime}}\leftarrow\theta^{\mu}$
 		\STATE Initialize replay buffer $\mathcal{R}\leftarrow\emptyset$
 		\FOR{ episode=1,M}
 		\STATE Initialize random process $\mathcal{N}$ for action exploration
 		\STATE Receive initial state observation $s_{1}$
 		\FOR{t=1,T}
 		\STATE Select action $a_{t} = \mu(s_{t}|\theta^{\mu})+\mathcal{N}_{t}$
 		\STATE Execute action $a_{t}$ and observe reward $r_{t}$ and new state $s_{t+1}$ 
 		\STATE Store transition $(s_{t},a_{t},r_{t},s_{t+1})$ in $R$
 		\STATE Sample minibatch of $N$ transitions $(s_{i},a_{i},r_{i},s_{i+1})$ from $R$
 		\IF{state $s_{i+1}$ is terminal state}
 		\STATE Set $y_{i}$ = $r_{i}$
 		\ELSE
 		\STATE Set $y_{i}$ = $r_{i}+\gamma Q^{\prime}(s_{i+1},\mu^{\prime}(s_{i+1}|\theta^{\mu^{\prime}})|\theta^{Q^{\prime}})$
 		\ENDIF
 		\STATE Update critic by minimizing loss:
 		\STATE $L=\frac{1}{N}\sum_{i}(y_{i}-Q(s_{i},a_{i}|\theta^{Q}))^{2}$
 		\STATE Update actor using sampled policy gradient:
 		\STATE $\nabla_{\theta^{\mu}}J\approx$
 		\STATE $\frac{1}{N}\sum_{i}\nabla_{a}Q(s_{i},a_{i}|\theta^{Q})|_{s=s_{i},a=\mu(s_{i})}\nabla_{\theta^{\mu}}\mu(s|\theta^{\mu})|_{s_{i}}$
 		\STATE Update target networks:
 		\STATE $\theta^{Q^{\prime}}\leftarrow\tau\theta^{Q}+(1-\tau)\theta^{Q^{\prime}}$,$\theta^{\mu^{\prime}}\leftarrow\tau\theta^{\mu}+(1-\tau)\theta^{\mu^{\prime}}$
 		\ENDFOR
 		\ENDFOR
 	\end{algorithmic}
 	\vspace{0em}
 \end{algorithm}
 \vspace{0em}

The algorithm we chose to use is the Deep Deterministic Policy Gradient (DDPG) algorithm \cite{cite:lillicrap2016ContinuousControl}, which is a model-free, off-policy RL algorithm based on Deterministic Policy Gradient  \cite{cite:silver2014DeterministicPolicy} and Deep Q Networks \cite{cite:mnih2015HumanLevel}. It is able to learn policies in high-dimensional, continuous state action spaces.

DDPG is a type of actor critic reinforcement learning algorithm, it uses two separate networks to parameterize the actor function and the critic function, respectively. The actor network $\mu(s|\theta^{\mu})$ maps the states to a deterministic action, and the critic network $Q(s,a|\theta^{Q})$ maps the state action pair to a Q-value.

The critic network is trained by minimizing the loss function:
\begin{equation}
\begin{aligned}
y_{t} &= r_{t}+\gamma Q^{\prime}(s_{t+1},a)|_{a=\mu^{\prime}(s_{t+1})} \\
L_{\text{Q}}(\theta^{Q}) &= \mathbb{E}_{s_{t},a_{t},r_{t},s_{t+1}\sim\mathcal{R}}\left[(Q(s_{t},a_{t})-y_{t})^2\right].
\end{aligned}
\label{eq:13}
\end{equation}

The actor network is trained by applying the deterministic policy gradient:%, which is the gradient of performance objective $J$ w.r.t the parameters of actor network $\theta^{\mu}$. 
\begin{equation}
\begin{aligned}
	\nabla_{\theta^{\mu}}J=\mathbb{E}_{s_{t}\sim\mathcal{R}}\left[\nabla_{a}Q(s,a|\theta^{Q})|_{s=s_{t},a=\mu(s_{t})}\nabla_{\theta^{\mu}}(s|\theta^{\mu})|_{s=s_{t}} \right].
\end{aligned}
\label{eq:14}
\end{equation}

%\textcolor{red}{In reinforcement learning , the objective is often to learn an optimal policy that maximizes the expected long tern cumulative reward, expressed as the discounted sum of immediate rewards with discount factor}

\subsection{Bounding Action Space}
\label{sec:3e}
The output of the network is desired joint angles, the rationale behind the selecting joint angles as the choice of action space is discussed in section \ref{sec:4}.

Joints have limits which restrict their range of movement, therefore we have to bound the action space within the angle limit. Using squashing sigmoid activation function such as $\tanh$ in the output unit is a common way in bounding network outputs. However, using $\tanh$ has its disadvantages: it would easily be saturated at the upper and lower bound of the range and require many updates to decrease, increasing the training time and hindering the performance. We use an approach proposed by Hausknecht et. al. called inverting gradients to bound the output action parameters  \cite{cite:hausknecht2015DeepReinforcement},
\begin{equation}
\nabla_{p}=\nabla_{p}\cdot\begin{cases}(p_{\text{max}}-p)/(p_{\text{max}}-p_{\text{min}}) & \text{if}~\nabla_{p}\geq0\\(p-p_{\text{min}})/(p_{\text{max}}-p_{\text{min}}) & \text{otherwise}\end{cases},
\label{eq:15}
\end{equation}
where $\nabla_{p}$ indicates the critic gradient with reference to action parameter, $p_{\text{max}}$, $p_{\text{min}}$, $p$ indicate the minimum, maximum and current activation of the action parameter, respectively.

From (\ref{eq:15}), we can see that with inverting gradients approach, the gradients are reduced as the output parameter approaches near the output boundary of the desired value range, and are inverted if the parameter exceeds the boundary, hereby restraining the output to the desired range.

%%%%%%%%%%%%%%%%%%%%%%%%%%%%%%%%%%%%%%%%%%%%%%%%%%%%%%%%%%%%%%%%%%%%%%%%%%%%%%%%

\section{Hierarchical Structure of High-level Learning and Low-level Control}
\label{sec:4}

\begin{figure}[t]
	\centering
	\vspace{0em}
	%\fbox{
	\includegraphics[width=0.8\linewidth, trim = 5.5cm 11cm 4cm 8.5cm, clip]{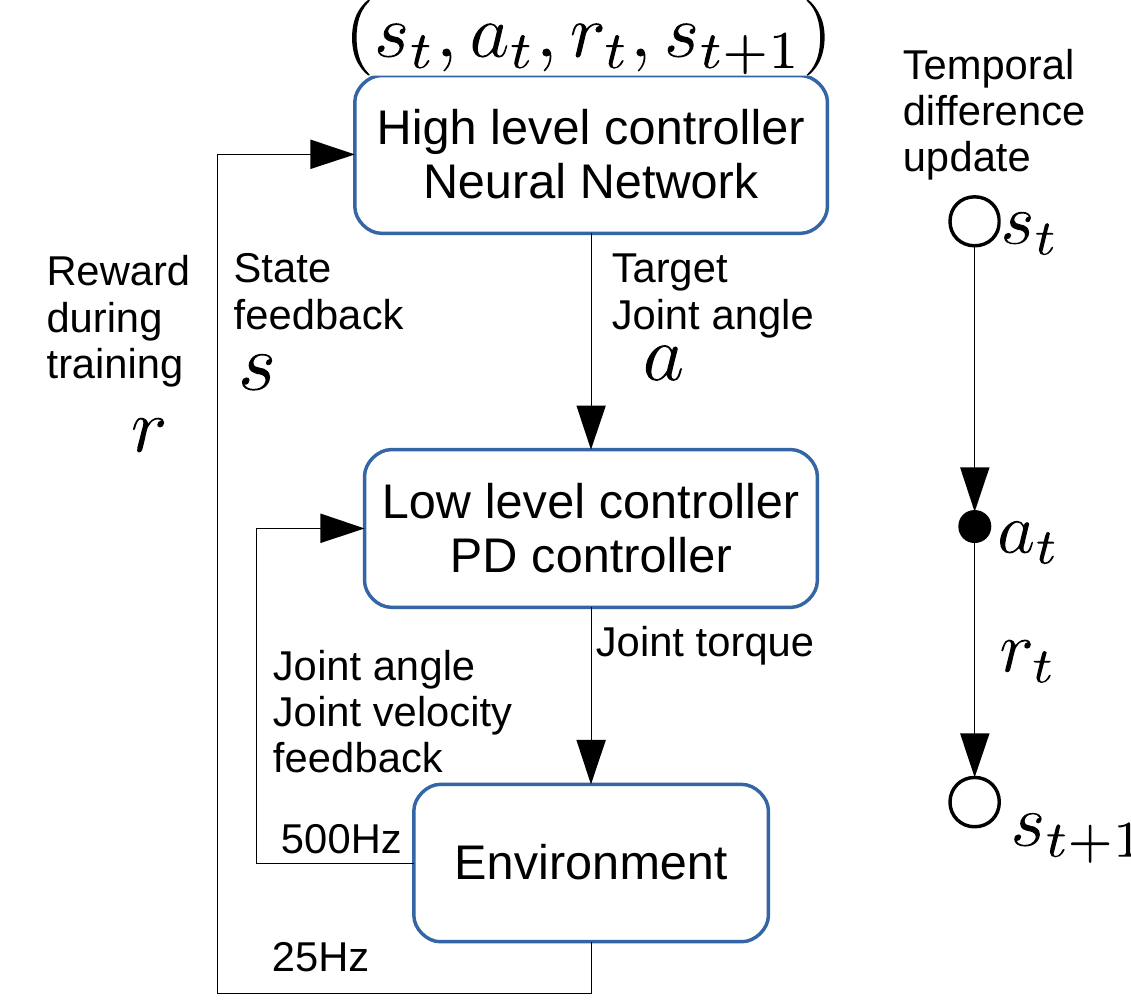}		
	%}
	\caption{System overview.}
	\label{fig:4}
	\vspace{0em}
\end{figure}

The idea of constructing the control architecture in a hierarchical manner is widely adopted by many studies \cite{cite:peng2017DeepLoco}, \cite{cite:heess2016LearningAnd}. In such hierarchical control systems, the Lower-level controller (LLC) and High-level controller (HLC) work at different frequencies, where usually the HLC works at a lower frequency.

The choice of output action parameterization has been proven to have a significant effect on the performance on reinforcement learning. Peng et al. compared the impact of four different actuation models that has different action parameterization on deep reinforcement learning: (1) direct torque control; (2) muscle activation for musculotendon units (MTU); (3) target joint angle for proportional-derivative controllers; (4) target joint angle velocity. Their study show that action parameterization including basic feedback such as target angle for PD control and muscle activation for MTU can improve policy performance and learning speed since such models are able to reflect the embodied biomechanical features more accurately  \cite{cite:peng2016LearningLocomotion}.

PD control has been proven to be a good action parameterization method as it is able to model the biomechanics of a system, and is easy to implement compared to other control methods such as MTU. Therefore, we choose joint angles as the output for the HLC learnt by DDPG and apply a PD controller as the LLC to translate the joint angle to torque for motor control. The overall structure of the control system is shown in Fig \ref{fig:4}.

\begin{figure}[t]
	\centering
	%\fbox{
	\includegraphics[width=0.95\linewidth, trim = {2.5cm 6.6cm 3.1cm 3.6cm}, clip]{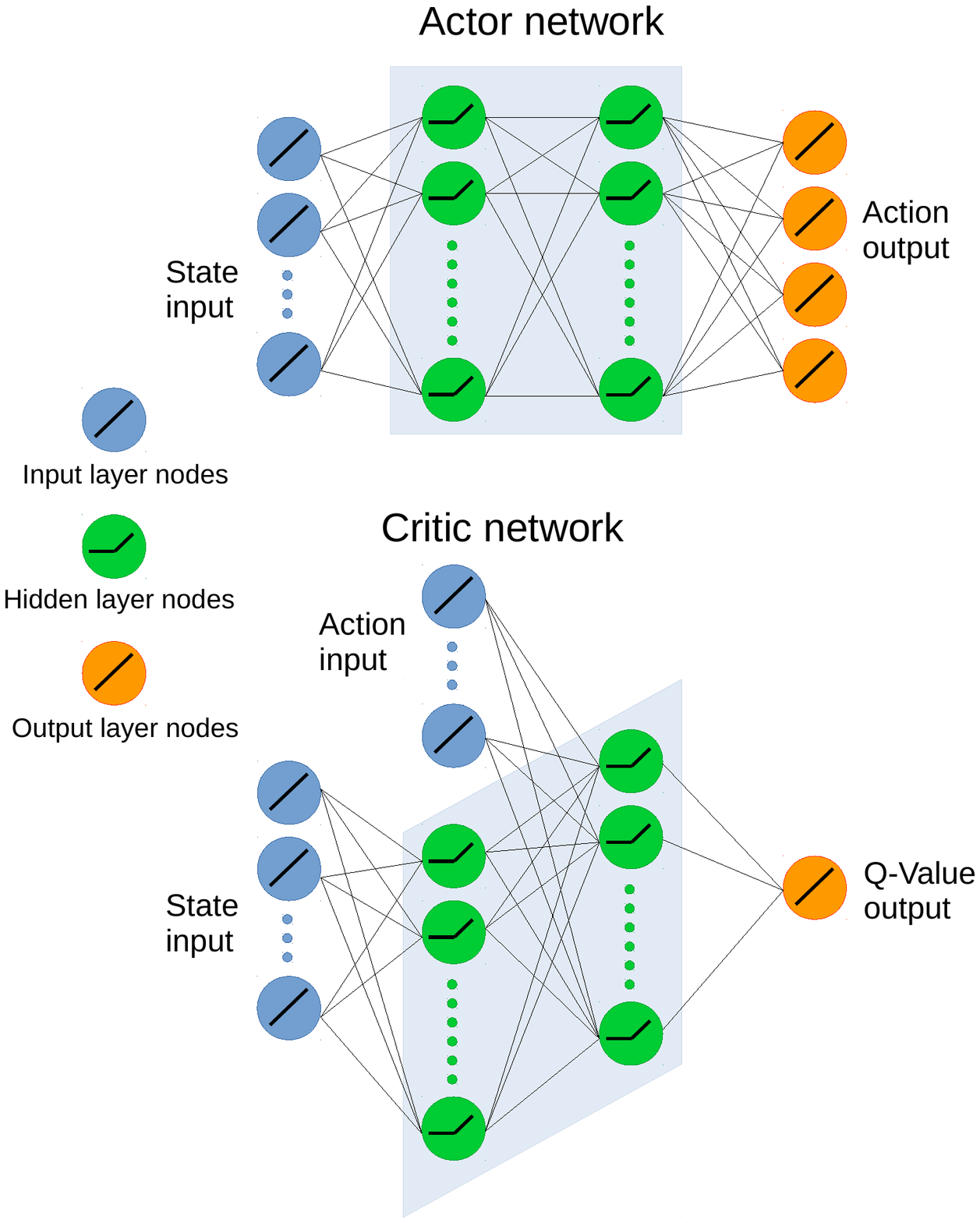}		
	%}
	\vspace{0em}
	\caption{Overview of neural network structure. }
	\label{fig:5}
	\vspace{0em}
\end{figure} 

\subsection{High-level Controller}
For the high level contoller we used DDPG to learn the control policy responsible for producing the desired motion synergies, i.e desired joint angles. The network structure is shown in Fig. \ref{fig:5}. Both the critic and actor network have 2 hidden layers, each hidden layer contains 100 nodes followed by a rectified linear unit (ReLU) activation function. In addition to the  state features, the critic network also takes into action parameters as input, the action value skips the first hidden layer and is directly forwarded to the second hidden layer. The output of the actor network is the 4 target joint angles. The network inputs consists of state features that are continuous, which are filtered through butterworth filters with a cutoff frequency of 10Hz, and discrete state features that are remained untouched.

%%%%%%%%%%%%%%%%%%%

\subsection{Low-level Controller}
We used PD controller as the low level controller, the input for the PD controller is desired joint angles produced by the HLC, and the output is joint torque. The feedback for the PD controller is filtered through a butterworth filter with a cutoff frequency of 50Hz. The parameters of the PD controller are shown in Table. \ref{tab:1}.

\begin{equation}
u=K_{p}(\theta_{\text{target}}-\theta_{\text{measured}})-K_{d}\dot{\theta}_{\text{measured}}
\label{15}
\end{equation}

\begin{table}[t]
	\centering
	\caption{PD controller parameters}
	\label{tab:1}	  
	\def\arraystretch{1.3}
	\begin{tabular}{p{2.0cm} p{1.0cm} p{1.0cm} p{1.0cm} p{1.0cm}}
		\hline
		& \multicolumn{4}{c}{Joints}\\
		\cline{2-5}
		PD parameters& Waist & Hip & Knee & Ankle\\
		\hline   
		$K_{p}$ (Nm/rad)  & 720 & 1080 & 2580 & 3160\\
		%\hline
		$K_{d}$ (Nms/rad)& 60 & 70 & 150 & 300\\
		%\hline
			
		\hline 
	\end{tabular}
\end{table}

%$K_{p}^{\text{Hip}} = 1080$ Nm/rad, $K_{d}^{\text{Hip}} = 70$ Nms/rad

%$K_{p}^{\text{Knee}} = 2580$ Nm/rad, $K_{d}^{\text{Knee}} = 150$ Nms/rad

%$K_{p}^{\text{Ankle}} = 3160$ Nm/rad, $K_{d}^{\text{Ankle}} = 300$ Nms/rad

%$K_{p}^{\text{Waist}} = 720$ Nm/rad, $K_{d}^{\text{Waist}} = 60$ Nms/rad

%%%%%%%%%%%

\begin{figure}[t]	
	\centering
	%\fbox{
	{\includegraphics[width=0.93\linewidth, trim = 4.0cm 13.1cm 5.0cm 12.7cm, clip]{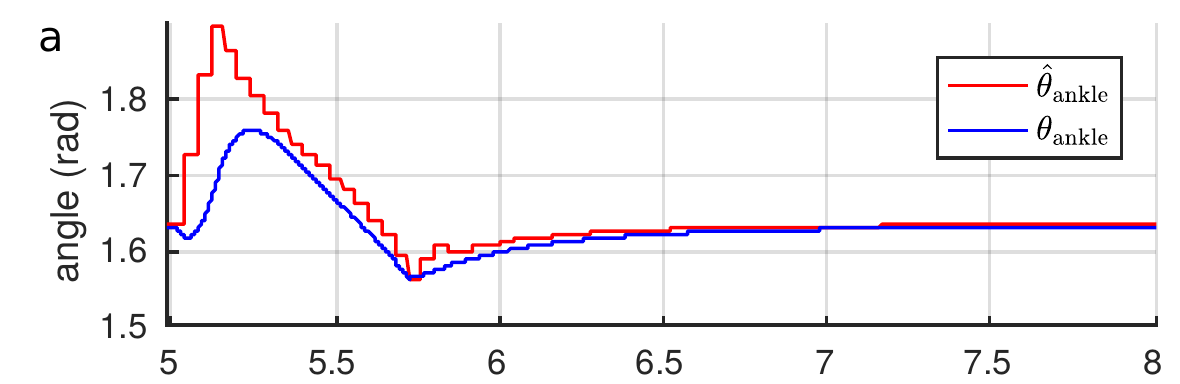}}
	%}\\
	%\vspace{0em}
	%\caption{Target and measured ankle joint anlgle.}
	\label{fig:5a}
	%\vspace{0em}
	
	%\fbox{
	{\includegraphics[width=0.93\linewidth, trim = 4.0cm 13.1cm 5.0cm 12.7cm, clip]{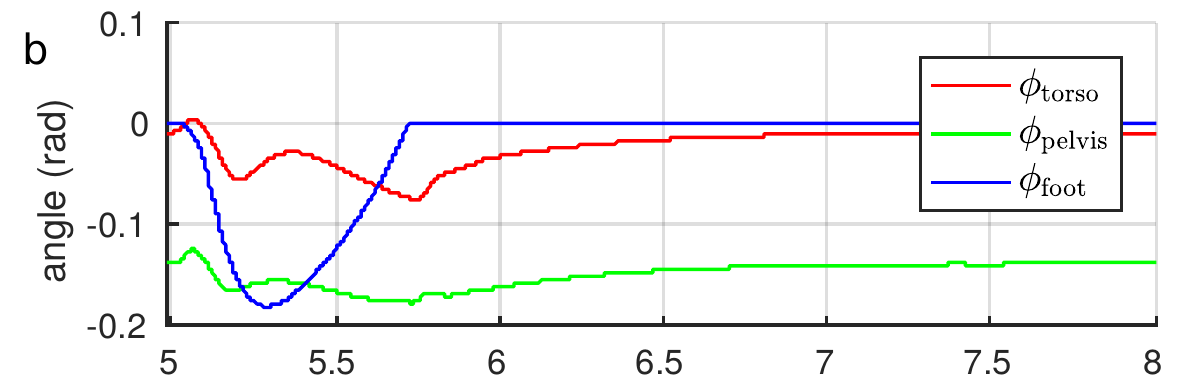}}
	%}\\
	%\vspace{0em}
	%\caption{Angular rate of the torso, pelvis and foot.}
	\label{fig:5b}
	%\vspace{0em}
	
	%\fbox{
	{\includegraphics[width=0.93\linewidth, trim = 4.0cm 13.1cm 5.0cm 12.7cm, clip]{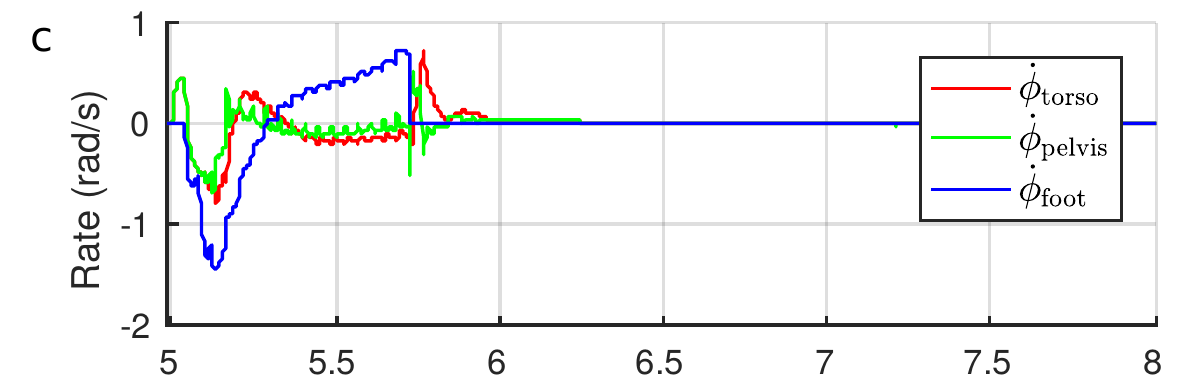}}
	%}\\
	%\vspace{0em}
	%\caption{Angular rate of the torso, pelvis and foot.}
	\label{fig:5c}
	%\vspace{0em}
	
	%\fbox{
	{\includegraphics[width=0.93\linewidth, trim = 4.0cm 13.1cm 5.0cm 12.7cm, clip]{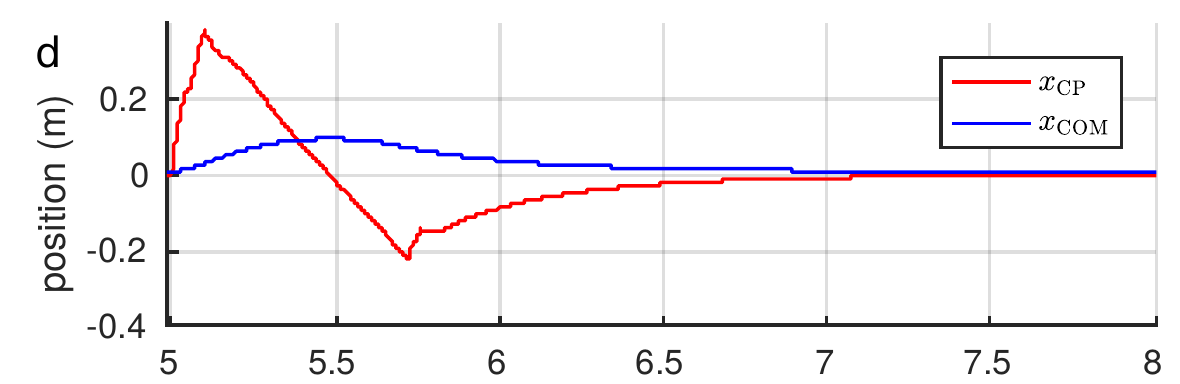}}
	%}\\
	%\vspace{0em}
	%\caption{Capture point and center of mass location.}
	\label{fig:5d}
	%\vspace{0em}
	
	%\fbox{
	{\includegraphics[width=0.93\linewidth, trim = 4.0cm 13.1cm 5.0cm 12.7cm, clip]{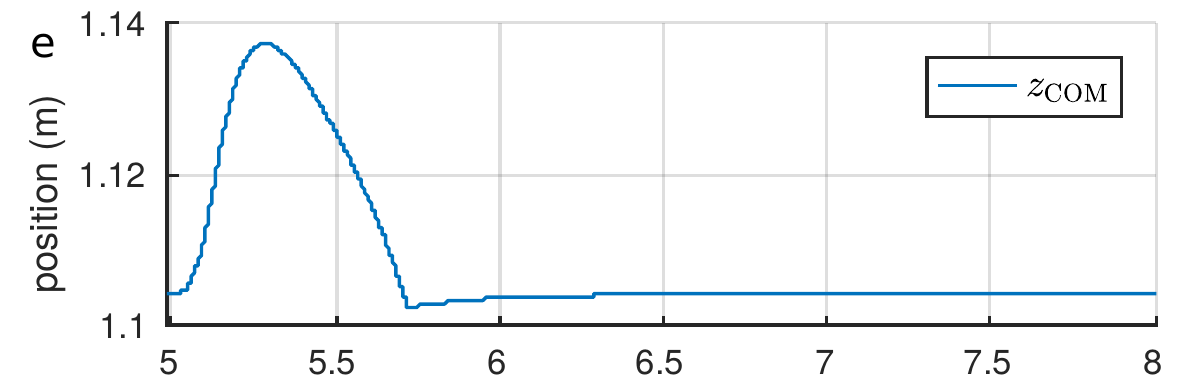}}
	%}\\
	%\vspace{0em}
	%\caption{Center of mass height}
	\label{fig:5e}
	%\vspace{0em}
	
	%\fbox{
	{\includegraphics[width=0.93\linewidth, trim = 4.0cm 12.6cm 5.0cm 12.7cm, clip]{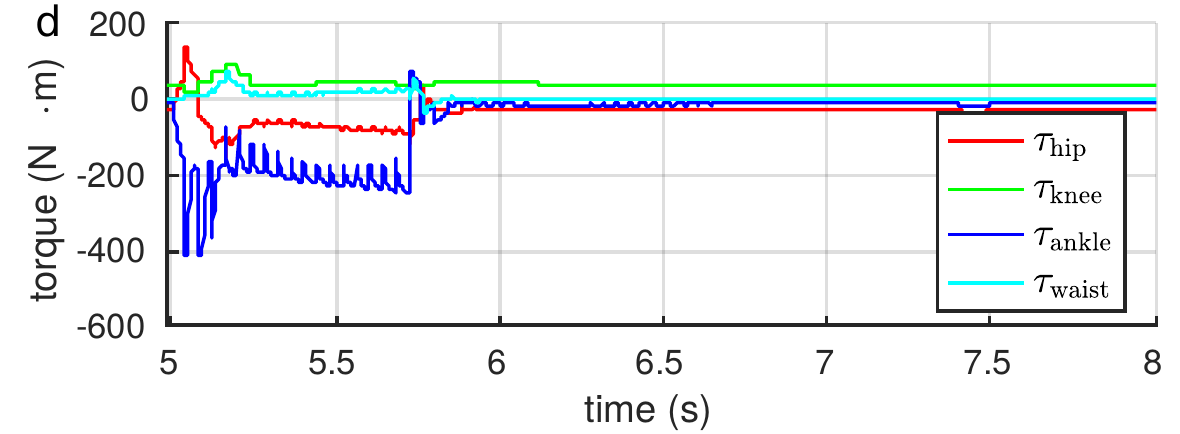}}
	%}\\
	\vspace{0em}
	\caption{Simulation data of forward push recovery (72.8 N$\cdot$s). (a) Reference and measured ankle joint angle; (b) Orientation of torso, pelvis and foot; (c) Angular rate of torso, pelvis and foot pitch; (d) Capture point and COM; (e) COM height; (f) Ankle joint torque.}
	\label{fig:5f}
	\vspace{0em}
	\label{fig:6}
\end{figure}

%%%%%%%%%%%%%%%%%%%%%%%%%%%%%%%%%%%%%%%%%%%%%%%%%%%%%%%%%%%%%%%%%%%%%%%%%%%%%%%%
\section{Learning Results}
\label{sec:5}

% % % % % % % % % % % % % % % % % % % % % %
\begin{figure}[t]
	\centering
	%\fbox{
		{\includegraphics[width=0.93\linewidth, trim = 4.0cm 13.1cm 5.0cm 12.7cm, clip]{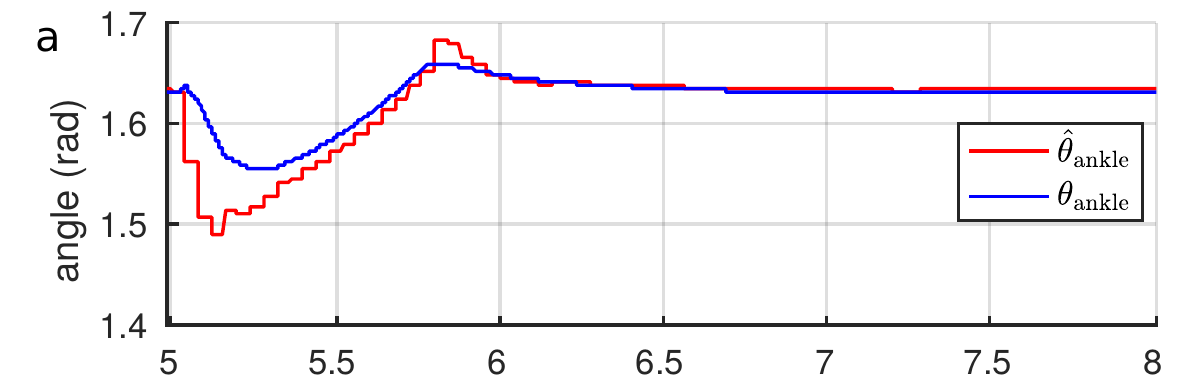}}
	%	}\\
	%\vspace{0em}
	%\caption{Target and measured ankle joint angle}
	\label{fig:6a}
	%\vspace{0em}

	%\fbox{
		{\includegraphics[width=0.93\linewidth, trim = 4.0cm 13.1cm 5.0cm 12.7cm, clip]{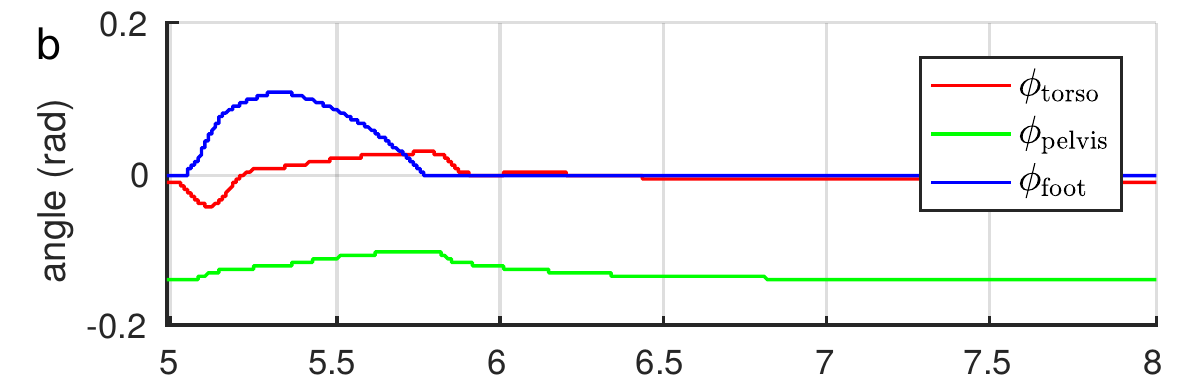}}
	%}\\
	%\vspace{0em}
	%\caption{Angular rate of the torso, pelvis and foot.}
	\label{fig:6b}
	%\vspace{0em}
	
	%\fbox{
		{\includegraphics[width=0.93\linewidth, trim = 4.0cm 13.1cm 5.0cm 12.7cm, clip]{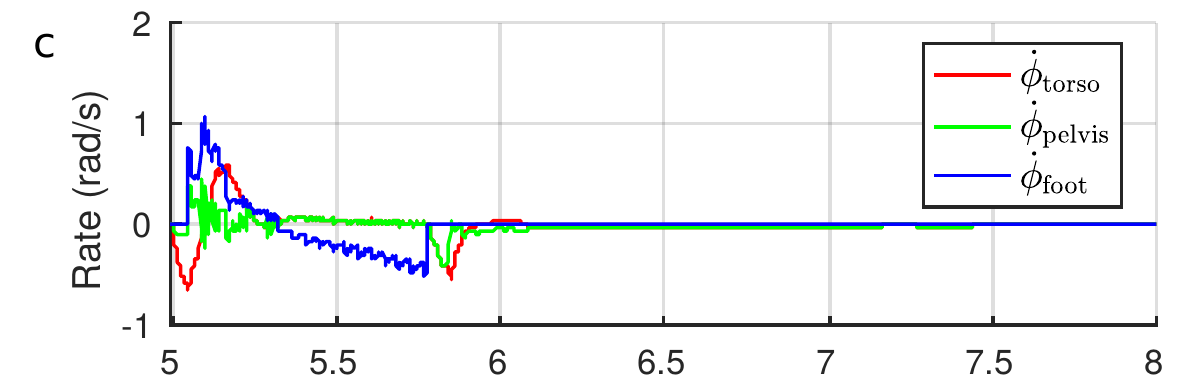}}
	%	}\\
	%\vspace{0em}
	%\caption{Angular rate of the torso, pelvis and foot.}
	\label{fig:6c}
	%\vspace{0em}
	
	%\fbox{
		{\includegraphics[width=0.93\linewidth, trim = 4.0cm 13.1cm 5.0cm 12.7cm, clip]{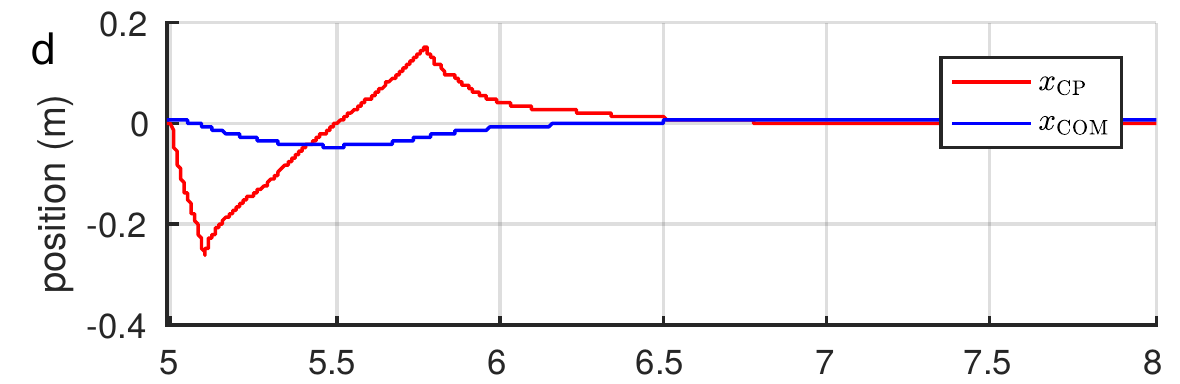}}
	%	}\\
	%\vspace{0em}
	%\caption{Capture point and center of mass location.}
	\label{fig:6d}
	%\vspace{0em}

	%\fbox{
		{\includegraphics[width=0.93\linewidth, trim = 4.0cm 13.1cm 5.0cm 12.7cm, clip]{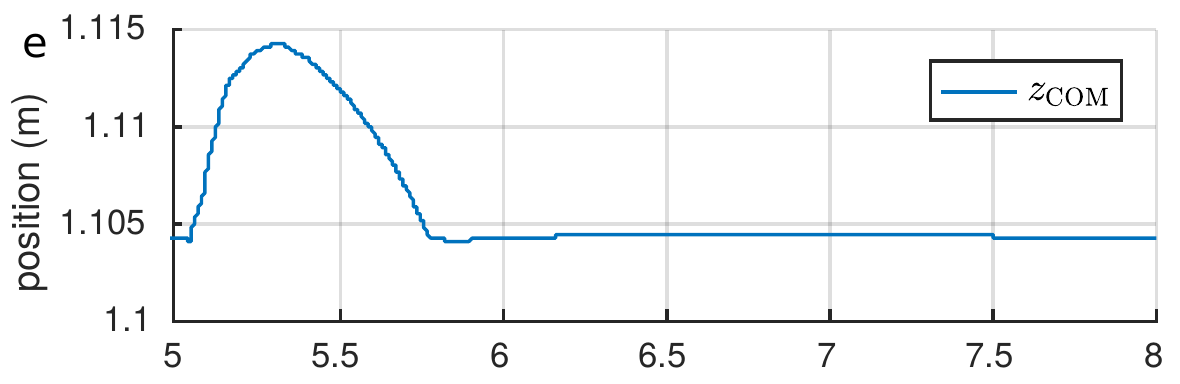}}
	%	}\\
	%\vspace{0em}
	%\caption{Center of mass height}
	\label{fig:6e}
	%\vspace{0em}
	
	%\fbox{
		{\includegraphics[width=0.93\linewidth, trim = 4.0cm 12.6cm 5.0cm 12.7cm, clip]{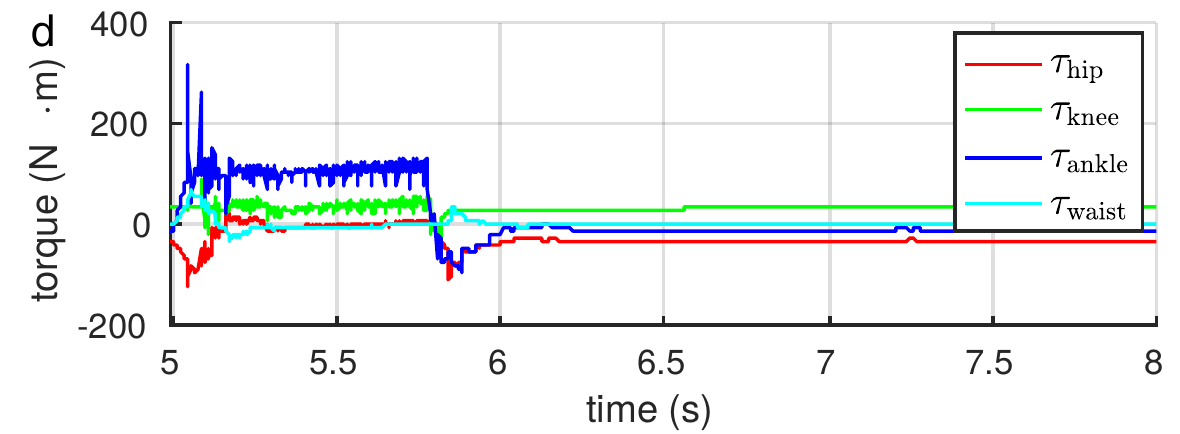}}
	%	}\\
	\vspace{0em}
	\caption{Simulation data of backward push recovery (-42.6 N$\cdot$s). (a) Reference and measured ankle joint angle; (b) Orientation of torso, pelvis and foot; (c) Angular rate of torso, pelvis and foot pitch; (d) Capture point and COM; (e) COM height; (f) Ankle joint torque.}
	\label{fig:6f}
	\vspace{0em}
	\label{fig:7}
\end{figure}

\begin{figure}[t] % I think this picture is ok actually
	\centering
	%\fbox{% figure is rotated width is actually height
	\includegraphics[width=0.95\linewidth, trim = 1.1cm 9.75cm 2.2cm 5.5cm, clip]{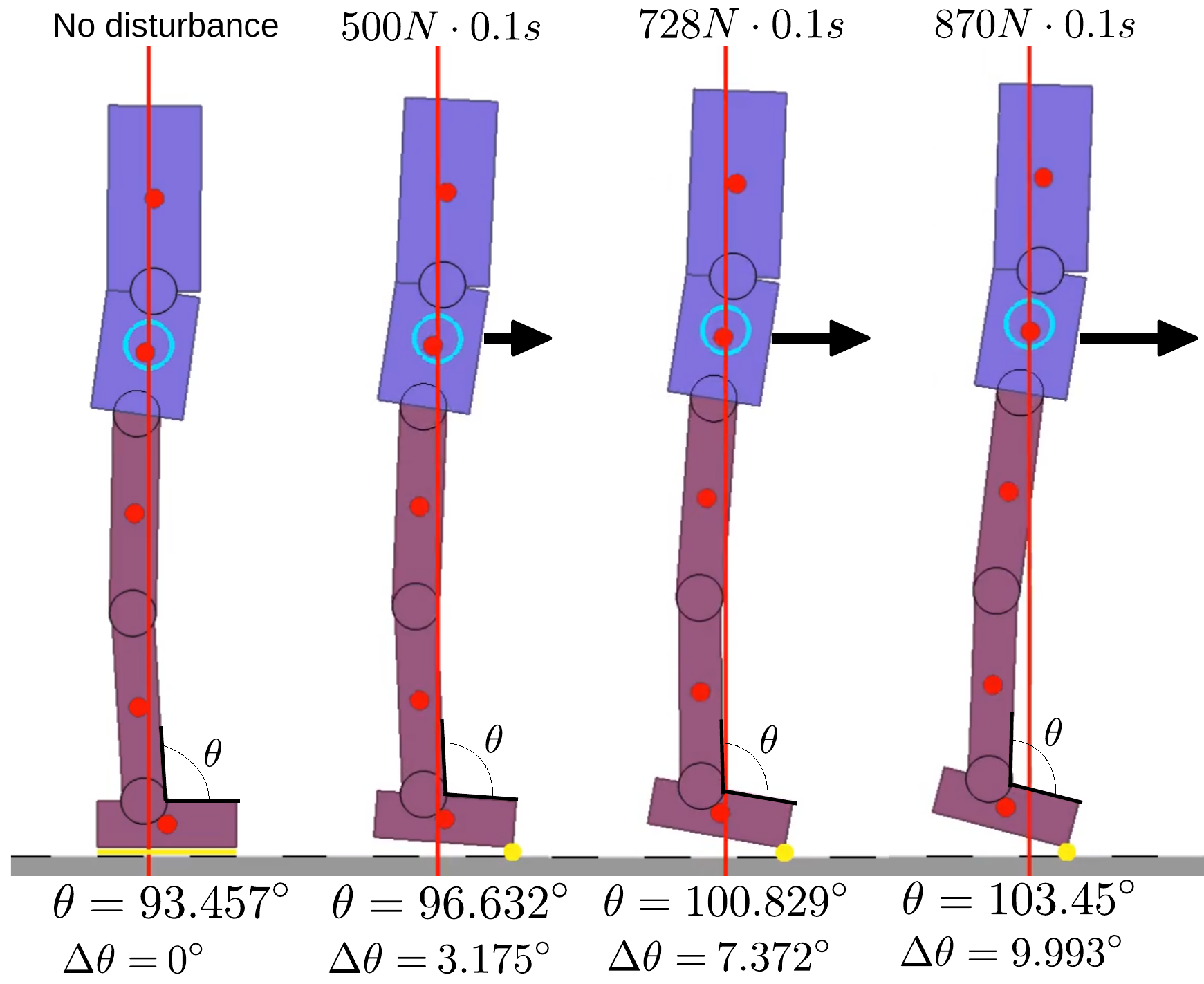}
	%}\\
	\caption{Maximum ankle angles generated by the policy and the change of angle w.r.t home position (Forward push).}
	\label{fig:8}
	\centering
	\vspace{0em}
\end{figure}

\begin{figure}[t] % I think this picture is ok actually
	\centering
	%\fbox{% figure is rotated width is actually height
	\includegraphics[width=0.95\linewidth, trim = 1.5cm 8.cm 1.8cm 7.25cm, clip]{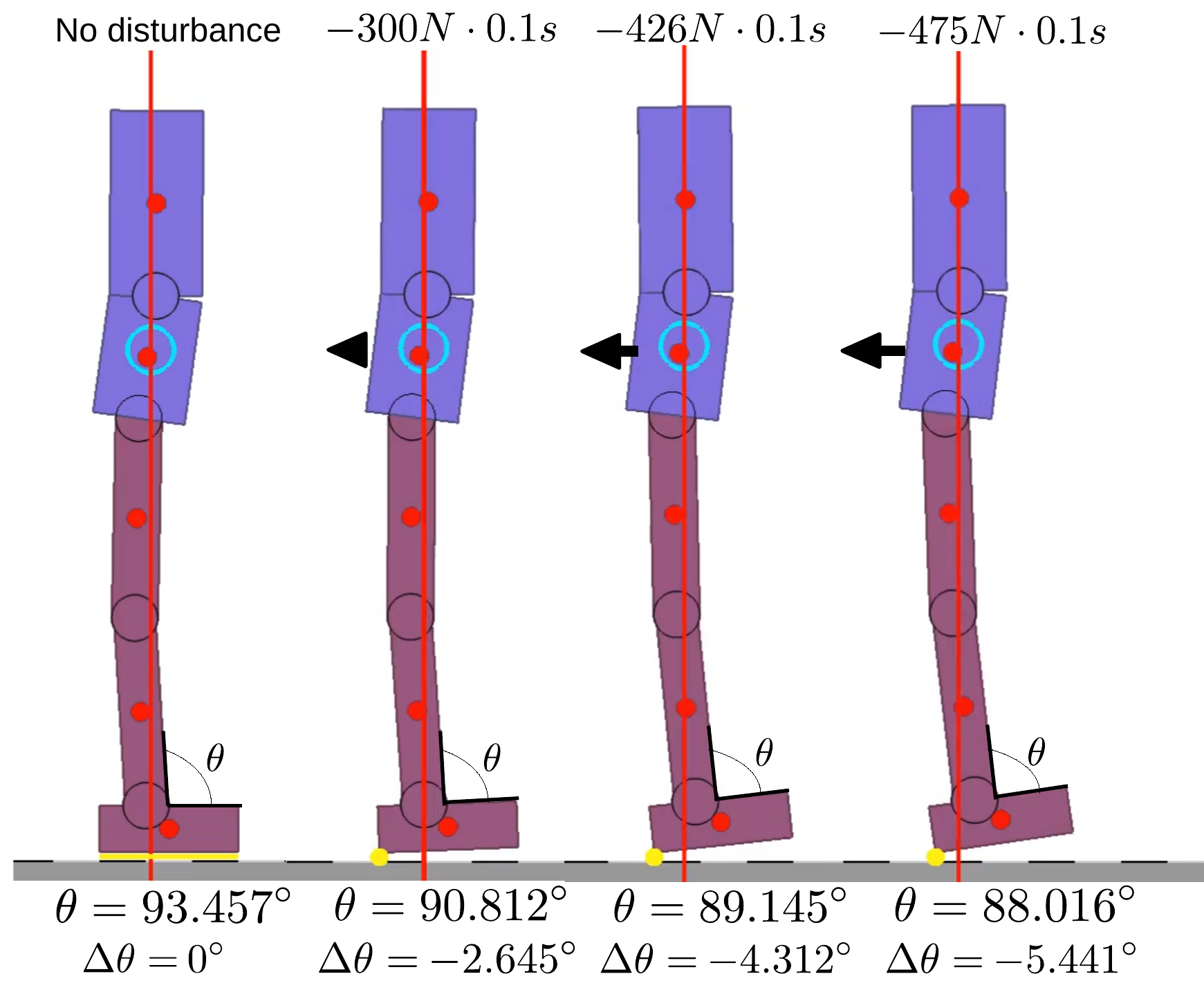}
	%}\\
	\caption{Maximum ankle angles generated by the policy and the change of angle w.r.t home position (Backward push).}
	\label{fig:9}
	\centering
	\vspace{0em}
\end{figure}

%\begin{figure*}[t] % I think this picture is ok actually
%	\centering
%	%\fbox{% figure is rotated width is actually height
%		\includegraphics[width=0.8\columnwidth, angle =-90, trim = 3.2cm 0.6cm 5cm 0.7cm, clip]{push}
%	%}\\
%	\caption{Maximum ankle angles generated by the policy under different amount of disturbances.}
%	\label{fig:8}
%	\centering
%	\vspace{0em}
%\end{figure*}

We can calculate the maximum rejectable impulse when the humanoid is in its stable balancing configuration according to capture point theory. In the stable configuration of the policy representation of the trained network, the horizontal distance from the COM to the front tip and back tip of the feet is respectively 0.189m and 0.111m. The height of the COM is 1.084m. According to (\ref{eq:1}), the maximum forward rejectable impulse is $72.8$ N$\cdot$s, maximum backward rejectable impulse is $42.6$ N$\cdot$s. A push force with duration of 0.1s is applied on the pelvis for simulating the impulse disturbance, therefore magnitude of the force are 728N and 426N respectively for the forward and backward pushes. Previous work has proven that foot tilting balancing strategy is capable of working under boundary rejectable impulse conditions \cite{cite:li2017HumanoidBalancing}. 

Fig. \ref{fig:6} and \ref{fig:7} respectively presents the data from forward and backward push recoveries. Fig. \ref{fig:8} and \ref{fig:9} show the snapshots of maximum ankle joint angles from successful balance recoveries of forward and backward pushes under various pushes. The snapshots show that under different amount of disturbances, the tilting angles of the foot and the angle of ankle joint are different. From the simulation results, it is observed that emerged behaviours are comparable to that of humans after sufficient amount of training without any prior knowledge explicitly given by designers:
\begin{itemize}
	\item Knee lock behavior naturally emerges;
	\item Heel/toe tipping behaviors naturally emerge;
	\item Able to work beyond the maximum rejectable impulse calculated using capture point (due to vertical motion);
	\item Able to actively push-off ankle joint and create foot tilting stably in response to different disturbance;
	\item Exploitation of maximum achievable ankle torque.
\end{itemize}

The amount of impulse the control system is capable of withstanding is slightly larger than the rejectable impulse calculated from the capture point theory, since the capture point uses a linear model assuming constant COM height. The network learns a balancing policy capable of withstanding impulse up to 87N$\cdot$s and -47.5N$\cdot$s, which is respectively 119.5\% and 111.5\% the amount of the forward and backward maximum rejectable impulse calculated using capture point (72.8N$\cdot$s and -42.6N$\cdot $s). This is because while the humanoid is pivoting around its toe, the horizontal velocity is partially redirected upward, increasing the height of the COM, therefore converting part of the kinetic energy into potential energy, slowing down the overall COM velocity. Learning-based control system is less restricted than traditional methods using ZMP in this aspect, because any stable action than improves the balance recovery will be reinforced such as ankle push-off, knee lock or any possible upper body movement. 

Some human-comparable features naturally emerge after sufficient amount of training without any prior knowledge given explicitly by humans. From Fig. \ref{fig:8} and \ref{fig:9}, we can see the policy actively increases the ankle angle to produce ankle push off behaviour. It is also shown that the humanoid has also learned to actuate its knee in a knee-lock configuration that minimizes knee torque and provides a lot of stability by simply exploiting the biomechanical constraint of the knee joint, very similar to what humans do.

The balance strategy learned by reinforcement learning has shown to have the ability to actively adjust the ankle joint angle and the tilting angle of the foot in response to the amount of disturbance applied. The active change in magnitude of ankle rotation $\Delta\theta$ relative to home position increases as the magnitude of force increases as seen in Fig. \ref{fig:8} and \ref{fig:9}.

Fig. \ref{fig:6}(f) and \ref{fig:7}(f) show the torque responses of all sagittal joints, where ankle joint in particular fully exploits the maximum achievable ankle torque for balance recovery. The control system responses to the disturbance by quickly generating ankle torque as large as possible, firstly larger than the gravitational torque for a short period to accelerate foot for tilting around the toe/heel, and then sustaining the maximum achievable torque for staying at the toe/heel with a total underactuation time about 0.8s. From the thickness, length and tilting angle of the foot, plus the mass of the body, it can be calculated that the magnitude of the maximum achievable torque while tilting around the toe and heel is respectively 216.14N$\cdot$m and 106.86N$\cdot$m. Simulation results from Fig. \ref{fig:6}(f) and \ref{fig:7}(f) shows that the magnitude of ankle torque applied during underactuation is around 210.41N$\cdot$m and 110.24N$\cdot$m for forward and backward push, which is close to the theoretical maximum achievable torque. The frontal section of the foot is longer than the rear section, thus the maximum achievable torque during forward push is larger than that of the backward push. %This is consistent with the maximum impulse that can be withstood, i.e. 87N$\cdot$s and -47.5N$\cdot$s for forward and backward push respectively. 

\section{Conclusion}
\label{sec:6}
Previous studies have already demonstrated that human-comparable balancing behaviours such as foot tilting behaviours can be achieved using deterministic and analytical engineering approaches. Our study in this paper concerns about whether it is possible to produce similar or better humanoid balance strategies that involves stable underactuated ankle push-off behaviour comparable to humans using deep reinforcement learning approach.

Our results demonstrated the feasibility and realizability of using deep reinforcement learning to learn a human-like balancing behavior with limited amount of prior structure being imposed on the control policy. We transfer knowledge from control engineering based methods and applied them into the design of rewards for RL. The importance of a physic based reward design shall be acknowledged. Otherwise it is difficult to balance the influence among different physical quantities and the balance behavior is difficult to be guaranteed by RL. The ankle push-off behaviour learned by RL is able to work robustly under circumstances where impulses are as much as the theoretical maximum that can be rejected. Moreover, deep RL has learned an adaptive way of actively changing ankle push-off angle in response to the applied disturbance.

The scope of this paper currently only covers standing balance in the sagittal plane in a 2D simulation as a proof-of-concept using learning approach. For future work, we plan to perform simulation in a 3D environment, and eventually, to apply learning based control to the real Valkyrie robot.

%%%%%%%%%%%%%%%%%%%%%%%%%%%%%%%%%%%%%%%%%%%%%%%%%%%%%%%%%%%%%%%%%%%%%%%%%%%%%%%%

%\begin{thebibliography}{99}
%\end{thebibliography}

\bibliographystyle{IEEEtran}
\bibliography{RAL-ICRA2017}

\begin{thebibliography}{10}
\providecommand{\url}[1]{#1}
\csname url@rmstyle\endcsname
\providecommand{\newblock}{\relax}
\providecommand{\bibinfo}[2]{#2}
\providecommand\BIBentrySTDinterwordspacing{\spaceskip=0pt\relax}
\providecommand\BIBentryALTinterwordstretchfactor{4}
\providecommand\BIBentryALTinterwordspacing{\spaceskip=\fontdimen2\font plus
\BIBentryALTinterwordstretchfactor\fontdimen3\font minus
  \fontdimen4\font\relax}
\providecommand\BIBforeignlanguage[2]{{%
\expandafter\ifx\csname l@#1\endcsname\relax
\typeout{** WARNING: IEEEtran.bst: No hyphenation pattern has been}%
\typeout{** loaded for the language `#1'. Using the pattern for}%
\typeout{** the default language instead.}%
\else
\language=\csname l@#1\endcsname
\fi
#2}}

\bibitem{cite:adamczyk2006AdvantagesRollingFoot}
P.~G. Adamczyk, S.~H. Collins, and A.~D. Kuo, ``The advantages of a rolling
  foot in human walking,'' \emph{Journal of Experimental Biology}, vol. 209,
  no.~20, pp. 3953--3963, 2006.

\bibitem{cite:li2015ActiveControl}
Z.~Li, C.~Zhou, Q.~Zhu, R.~Xiong, N.~Tsagarakis, and D.~Caldwell, ``Active
  control of under-actuated foot tilting for humanoid push recovery,'' in
  \emph{Proc. IEEE/RSJ Int. Conf. Intell. Robots and Syst.}, 2015, pp.
  977--982.

\bibitem{cite:li2017HumanoidBalancing}
Z.~Li, C.~Zhou, Q.~Zhu, and R.~Xiong, ``Humanoid balancing behavior featured by
  underactuated foot motion,'' \emph{IEEE Transactions on Robotics}, vol.~33,
  no.~2, pp. 298--312, 2017.

\bibitem{cite:heess2017EmergenceOf}
N.~Heess, S.~Sriram, J.~Lemmon, J.~Merel, G.~Wayne, Y.~Tassa, T.~Erez, Z.~Wang,
  A.~Eslami, M.~Riedmiller, \emph{et~al.}, ``Emergence of locomotion behaviours
  in rich environments,'' \emph{preprint arXiv:1707.02286}, 2017.

\bibitem{cite:schulman2015TrustRegion}
J.~Schulman, S.~Levine, P.~Abbeel, M.~Jordan, and P.~Moritz, ``Trust region
  policy optimization,'' in \emph{Proc. Int. Conf. Machine Learning}, 2015, pp.
  1889--1897.

\bibitem{cite:gu2016ContinuousDeepQ}
S.~Gu, T.~Lillicrap, I.~Sutskever, and S.~Levine, ``Continuous deep q-learning
  with model-based acceleration,'' in \emph{Proc. Int. Conf. Machine Learning},
  2016, pp. 2829--2838.

\bibitem{cite:mnih2016AsynchronousMethods}
V.~Mnih, A.~P. Badia, M.~Mirza, A.~Graves, T.~Lillicrap, T.~Harley, D.~Silver,
  and K.~Kavukcuoglu, ``Asynchronous methods for deep reinforcement learning,''
  in \emph{Proc. Int. Conf. Machine Learning}, 2016, pp. 1928--1937.

\bibitem{cite:lillicrap2016ContinuousControl}
T.~P. Lillicrap, J.~J. Hunt, A.~Pritzel, N.~Heess, T.~Erez, Y.~Tassa,
  D.~Silver, and D.~Wierstra, ``Continuous control with deep reinforcement
  learning,'' \emph{preprint arXiv:1509.02971}, 2015.

\bibitem{cite:van2007ReinforcementLearning}
H.~Van~Hasselt and M.~A. Wiering, ``Reinforcement learning in continuous action
  spaces,'' in \emph{IEEE Int. Symp. Approximate Dynamic Programming and
  Reinforcement Learning}, 2007, pp. 272--279.

\bibitem{cite:van2012ReinforcementLearning}
H.~Van~Hasselt, ``Reinforcement learning in continuous state and action
  spaces,'' in \emph{Reinforcement Learning}, 2012, pp. 207--251.

\bibitem{cite:peng2015DynamicTerrain}
X.~B. Peng, G.~Berseth, and M.~Van~de Panne, ``Dynamic terrain traversal skills
  using reinforcement learning,'' \emph{ACM Transactions on Graphics}, vol.~34,
  no.~4, p.~80, 2015.

\bibitem{cite:peng2017DeepLoco}
X.~B. Peng, G.~Berseth, K.~Yin, and M.~van~de Panne, ``Deeploco: Dynamic
  locomotion skills using hierarchical deep reinforcement learning,'' \emph{ACM
  Transactions on Graphics}, vol.~36, no.~4, 2017.

\bibitem{cite:peng2016TerrainAdaptive}
X.~B. Peng, G.~Berseth, and M.~Van~de Panne, ``Terrain-adaptive locomotion
  skills using deep reinforcement learning,'' \emph{ACM Transactions on
  Graphics}, vol.~35, no.~4, p.~81, 2016.

\bibitem{cite:hyon2009IntegrationOf}
S.-H. Hyon, R.~Osu, and Y.~Otaka, ``Integration of multi-level postural
  balancing on humanoid robots,'' in \emph{Proc. IEEE Int. Conf. Robot. Autom.,
  2009. ICRA'09.}, 2009, pp. 1549--1556.

\bibitem{cite:stephens2010DynamicBalance}
B.~J. Stephens and C.~G. Atkeson, ``Dynamic balance force control for compliant
  humanoid robots,'' in \emph{Proc. IEEE/RSJ Int. Conf. Intell. Robots and
  Syst.}, 2010, pp. 1248--1255.

\bibitem{cite:li2012StabilizationFor}
Z.~Li, B.~Vanderborght, N.~G. Tsagarakis, L.~Colasanto, and D.~G. Caldwell,
  ``{Stabilization for the compliant humanoid robot COMAN exploiting intrinsic
  and controlled compliance}{\tiny },'' in \emph{Proc. IEEE Int. Conf. Robot.
  Autom.}, 2012, pp. 2000--2006.

\bibitem{cite:pratt2006CapturePoint}
J.~Pratt, J.~Carff, S.~Drakunov, and A.~Goswami, ``Capture point: A step toward
  humanoid push recovery,'' in \emph{Proc. IEEE-RAS Int. Conf. Humanoid
  Robots}, 2006, pp. 200--207.

\bibitem{cite:amodei2016ConcreteProblems}
D.~Amodei, C.~Olah, J.~Steinhardt, P.~Christiano, J.~Schulman, and D.~Man{\'e},
  ``{Concrete problems in AI safety},'' \emph{arXiv:1606.06565}, 2016.

\bibitem{cite:ng2000AlgorithmsFor}
A.~Y. Ng, S.~J. Russell, \emph{et~al.}, ``Algorithms for inverse reinforcement
  learning.'' in \emph{Proc. Int. Conf. Machine learning}, 2000, pp. 663--670.

\bibitem{cite:abbeel2004ApprenticeshipLearning}
P.~Abbeel and A.~Y. Ng, ``Apprenticeship learning via inverse reinforcement
  learning,'' in \emph{Proc. Int. Conf. Machine learning}, 2004, p.~1.

\bibitem{cite:christiano2017DeepReinforcement}
P.~Christiano, J.~Leike, T.~B. Brown, M.~Martic, S.~Legg, and D.~Amodei, ``Deep
  reinforcement learning from human preferences,'' \emph{preprint
  arXiv:1706.03741}, 2017.

\bibitem{cite:silver2014DeterministicPolicy}
D.~Silver, G.~Lever, N.~Heess, T.~Degris, D.~Wierstra, and M.~Riedmiller,
  ``Deterministic policy gradient algorithms,'' in \emph{Proc. Int. Conf.
  Machine Learning}, 2014, pp. 387--395.

\bibitem{cite:mnih2015HumanLevel}
V.~Mnih, K.~Kavukcuoglu, D.~Silver, A.~A. Rusu, J.~Veness, M.~G. Bellemare,
  A.~Graves, M.~Riedmiller, A.~K. Fidjeland, G.~Ostrovski, \emph{et~al.},
  ``Human-level control through deep reinforcement learning,'' \emph{Nature},
  vol. 518, no. 7540, pp. 529--533, 2015.

\bibitem{cite:hausknecht2015DeepReinforcement}
M.~Hausknecht and P.~Stone, ``Deep reinforcement learning in parameterized
  action space,'' \emph{preprint arXiv:1511.04143}, 2015.

\bibitem{cite:heess2016LearningAnd}
N.~Heess, G.~Wayne, Y.~Tassa, T.~Lillicrap, M.~Riedmiller, and D.~Silver,
  ``Learning and transfer of modulated locomotor controllers,'' \emph{preprint
  arXiv:1610.05182}, 2016.

\bibitem{cite:peng2016LearningLocomotion}
X.~B. Peng and M.~van~de Panne, ``Learning locomotion skills using deeprl: does
  the choice of action space matter?'' in \emph{Proc. the ACM
  SIGGRAPH/Eurographics Symp. on Computer Animation}, 2017, p.~12.

\end{thebibliography}
%\bibliography{backup.bib}

\end{document}